\documentclass[11pt,a4paper]{article}
\usepackage{CJKutf8}
\usepackage[hyperref]{acl2021}
\usepackage{times}
\usepackage{latexsym}
\usepackage{amsmath}
\usepackage{tikz}
\usepackage{color}
\usepackage[svgnames]{xcolor}
\usepackage{hyperref}
\usepackage{bbm}
\usepackage{multirow}
\usepackage{float}
\usepackage{mathtools}
\usepackage{placeins}
\usepackage{graphicx}
\usepackage{tabularx}
\usepackage{xspace,mfirstuc,tabulary}

\colorlet{LightRed}{Red!35!White}

\newcommand{\ignore}[1]{}

\newcommand{\model}[3]{\textit{#1}_\textit{#2}\textit{(#3)}}
\newenvironment{itemizesquish}{\begin{list}{\setcounter{enumi}{0}\labelitemi}{\setlength{\itemsep}{-0.25em}\setlength{\labelwidth}{0.5em}\setlength{\leftmargin}{\labelwidth}\addtolength{\leftmargin}{\labelsep}}}{\end{list}}

\urldef\tfdlink\url{https://www.tensorflow.org/datasets/catalog/c4#c4multilingual}

\usepackage{microtype}

\aclfinalcopy % Uncomment this line for the final submission
\def\aclpaperid{2887} %  Enter the acl Paper ID here

\title{Diverse Pretrained Context Encodings Improve Document Translation}

\author{Domenic Donato, Lei Yu, Chris Dyer \\
  DeepMind \\
  London, United Kingdom \\
  \texttt{\{domenicd,leiyu,cdyer\}@deepmind.com}}

\date{}

\begin{document}
\maketitle
\begin{abstract}
We propose a new architecture for adapting a sentence-level sequence-to-sequence transformer by incorporating multiple pretrained document context signals and assess the impact on translation performance of (1)~different pretraining approaches for generating these signals, (2)~the quantity of parallel data for which document context is available, and (3)~conditioning on source, target, or source and target contexts. Experiments on the NIST Chinese--English, and IWSLT and WMT English--German tasks support four general conclusions: that using pretrained context representations markedly improves sample efficiency, that adequate parallel data resources are crucial for learning to use document context, that jointly conditioning on multiple context representations outperforms any single representation, and that source context is more valuable for translation performance than target side context. Our best multi-context model consistently outperforms the best existing context-aware transformers.
\end{abstract}

\section{Introduction}

Generating an adequate translation for a sentence often requires understanding the context in which the sentence occurs (and in which its translation will occur). Although single-sentence translation models demonstrate remarkable performance~\citep{chen-etal-2018-best,VaswaniSPUJGKP17,bahdanau:2015}, extra-sentential information can be necessary to make correct decisions about lexical choice, tense, pronominal usage, and stylistic features, and therefore designing models capable of using this information is a necessary step towards fully automatic high-quality translation. A series of papers have developed architectures that permit the broader translation model to condition on extra-sentential context~\citep{ZhangLSZXZL18,WerlenRPH18}, operating jointly on multiple sentences at once~\citep{Junczys-Dowmunt19}, or indirectly conditioning on target side document context using Bayes' rule~\citep{DBLP:journals/tacl/YuSSLKBD20}.

While noteworthy progress has been made at modeling monolingual documents~\citep{gpt-3}, progress on document translation has been less remarkable, and continues to be hampered by the limited quantities of parallel document data relative to the massive quantities of monolingual document data. One recurring strategy for dealing with this data scarcity---and which is the basis for this work---is to adapt a sentence-level sequence-to-sequence model by making additional document context available in a second stage of training~\citep{MarufMH19,ZhangLSZXZL18,WerlenRPH18,HaffariM18}. This two-stage training approach provides an inductive bias that encourages the learner to explain translation decisions preferentially in terms of the current sentence being translated, but these can be modulated at the margins by using document context. However, a weakness of this approach is that the conditional dependence of a translation on its surrounding context given the source sentence is weak, and learning good context representations purely on the basis of scarce parallel document data is challenging.

A recent strategy for making better use of document context in translation is to use pretrained BERT representations of the context, rather than learning them from scratch~\cite{bert-mt}. Our key architectural innovation in this paper is an architecture for two-staged training that enables jointly conditioning on \emph{multiple} context types, including both the source and target language context. Practically, we can construct a weak context representation from a variety of different contextual signals, and these are merged with the source sentence encoder's representation at each layer in the transformer. To examine the potential of this architecture, we explore two high-level research questions. First, using source language context, we explore the relative impact of different kinds of pretraining objectives on the performance obtained (BERT and PEGASUS), the amount of parallel document training data required, and the size of surrounding context. Second, recognizing that maintaining consistency in translation would seem to benefit from larger contexts in the target language, we compare the impact of source language context, target language context, and context containing both.

Our main findings are (1)~that multiple kinds of source language context improves performance of document translation over existing contextual representations, especially those that do not use pretrained context representations; (2)~that although fine-tuning using pretrained contextual representations improves performance, large performance is strongly determined by the availability of contextual parallel data; and (3)~that while both source and target language context provide benefit, source language context is more valuable, unless the quality of the target language context translations is extremely high.

\section{Model Description}
\label{sec:model}

Our architecture is designed to incorporate multiple sources of external embeddings into a pretrained sequence-to-sequence transformer model. We execute this by creating a new attention block for each embedding we wish to incorporate and stack them. We then insert this attention stack as a branching path in each layer of the encoder and decoder. The outputs of the new and original paths are averaged before being passed to the feed forward block at the end of the layer.  Details are discussed below (\S\ref{sec:parallel}), and the architecture is shown in Figure~\ref{fig:model_diagram}.

The model design follows the \textit{adapter pattern} \cite{designPatterns}. The interface between the external model and translation model takes the form of an attention block which learns to perform the adaptation. The independence between the models means that different input data can be provided to each, which enables extra information during the translation process. In this work, we leverage this technique to: (1)~enhance a sentence-level model with additional source embeddings; (2)~convert a sentence-level model to a document-level model by providing contextual embeddings. Like BERT-fused \cite{bert-mt}, we use pretrained masked language models to generate the external embeddings. 

\subsection{Pre-Trained Models}

We use two kinds of pretrained models: BERT~\citep{devlin-2019-bert} and PEGASUS~\citep{pegasus}. Although similar in architecture, we conjecture that these models will capture different signals on account of their different training objectives.

\paragraph{BERT} is trained with a masked word objective and a two sentence similarity classification task. During training, it is provided with two sentences that may or may not be adjacent, with some of their words masked or corrupted. BERT predicts the correct words and determining if the two sentences form a contiguous sequence. Intuitively, BERT provides rich word-in-context embeddings. In terms of machine translation, it's reasonable to postulate that BERT would provide superior representations of the source sentence and reasonable near sentence context modulation. On the other hand, we expect it to fail to provide contextual conditioning when the pair of sentences are not adjacent. This shortcoming is where PEGASUS comes in.

\paragraph{PEGASUS} is trained with a masked sentence objective. During training, it is given a document that has had random sentences replaced by a mask token. Its task is to decode the masked sentences in the same order they appear in the document. As a result, PEGASUS excels at summarization tasks, which require taking many sentences and compressing them into a representation from which another sentence can be generated. In terms of providing context for document translation, we conjecture that PEGASUS will be able to discover signals across longer ranges that modulate output.

\subsection{Embedding Notation}

To keep track of the type of embeddings being incorporated in a particular configuration, we use the notational convention $\model{Model}{Side}{Inputs}$.

\begin{itemizesquish}
    \item Model: $\textit{B}$ for BERT, $\textit{P}$ for PEGASUS, and $\textit{D}$ for Document Transformer \cite{ZhangLSZXZL18}.
    \item Side: $\textit{s}$ for the source and $\textit{t}$ for the target language.
    \item Inputs: $\textit{c}$ for the current source (or target), i.e., $\boldsymbol{x}_i$, $\textit{p}$ for the previous source (target), and $\textit{n}$ for the next one. Note that $\textit{3p}$ means \textit{the three previous} sources (targets), $(\boldsymbol{x}_{i - 3}, \boldsymbol{x}_{i - 2}, \boldsymbol{x}_{i - 1})$.
    \item When multiple embeddings are used, we include a $\Rightarrow$ to indicate the order of attention operations.
\end{itemizesquish}

\noindent We can thus represent the BERT-fused document model proposed by \citet{bert-mt} as $\model{B}{s}{p,c}$ since it passes the previous and current source sentences as input to BERT.

\subsection{Enhanced Models}

The core of this work is to understand the benefits that adding a diverse set of external embeddings has on the quality of document translation. To this effect, we introduce two new models that leverage the output from both BERT and PEGASUS:
\begin{alignat*}{2}
&\textit{Multi-source} &&\coloneqq \model{B}{s}{c} \Rightarrow \model{P}{s}{c}  \\
&\textit{Multi-context} &&\coloneqq \model{B}{s}{p,c} \Rightarrow \model{B}{s}{c,n} \Rightarrow \model{P}{s}{3p,c,3n}
\end{alignat*} 

\begin{figure*}[t]
\centering
\includegraphics[width=1.00\textwidth]{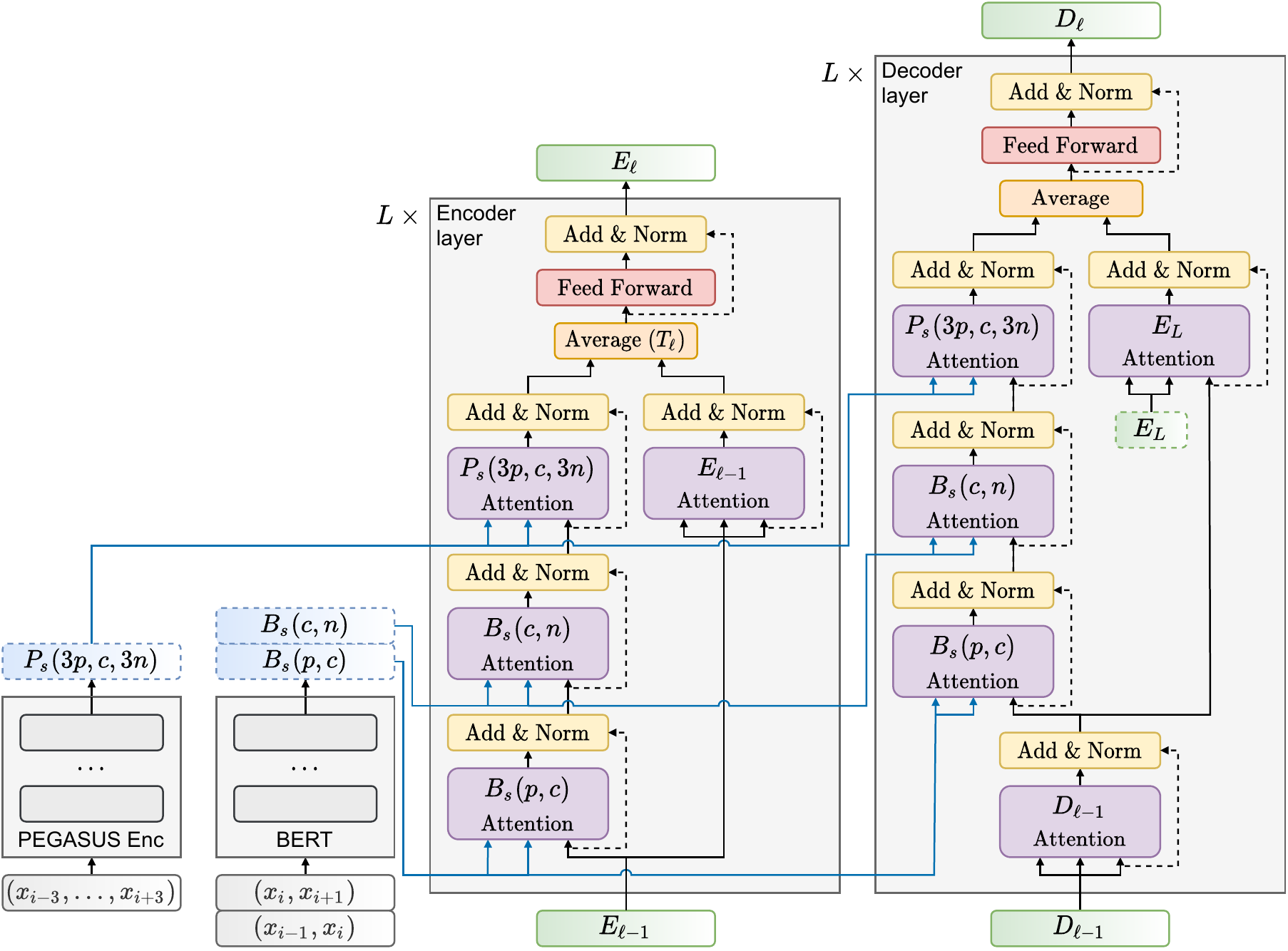}
\caption{Architecture of our Multi-context model. The pretrained PEGASUS Encoder and BERT model along with their inputs and resulting embeddings are shown on the left. In this configuration, a batch of two different sentence pairs are passed to BERT per translation example. The left to right ordering of the three inputs going into an attention block are: \textit{keys}, \textit{values}, \textit{queries}. During training the \textit{average} operation is replaced with \textit{drop-branch}. A partially shaded box indicates data while full shading is used for operations. A dashed border means the data is constant for a given translation. We use dashed arrows for residual connections and blue arrows to indicate embeddings that originate from outside the transformer model.}
\label{fig:model_diagram}
\end{figure*}

There are a few ways to integrate the output of external models into a transformer layer. We could stack them \textit{vertically} after the self-attention block \cite{ZhangLSZXZL18} or we could place them \textit{horizontally} and average all of their outputs together like MAT \cite{fan2020multibranch}. Our preliminary experiments show that the parallel attention stack, depicted in Figure \ref{fig:model_diagram}, works best. Therefore, we adopt this architecture in our experiments.

\subsection{Parallel Attention Stack}
\label{sec:parallel}
If we let $\mathbf{A} = \model{B}{s}{p,c}$, $\mathbf{B} = \model{B}{s}{c,n}$, and $\mathbf{C} = \model{P}{s}{3p,c,3n}$ refer to the output of the external pretrained models computed once per translation example, then the Multi-context encoder layer is defined as
\begin{alignat*}{2}
    &\mathbf{R}_\ell &&= \textit{AttnBlock}(\mathbf{E}_{\ell - 1},\ \mathbf{E}_{\ell - 1},\ \mathbf{E}_{\ell - 1}) \\
    &\mathbf{S}^a_\ell &&= \textit{AttnBlock}(\mathbf{A},\ \mathbf{A},\ \mathbf{E}_{\ell - 1}) \\
    &\mathbf{S}^b_\ell &&= \textit{AttnBlock}(\mathbf{B},\ \mathbf{B},\ \mathbf{S}^a_\ell) \\
    &\mathbf{S}_\ell &&= \textit{AttnBlock}(\mathbf{C},\ \mathbf{C},\ \mathbf{S}^b_\ell) \\
    &\mathbf{T}_\ell &&= \left\{
        \begin{array}{ll}
            \textit{DropBranch}(\mathbf{R}_\ell,\ \mathbf{S}_\ell) & \quad \textit{training} \\
            \frac{1}{2} \cdot (\mathbf{R}_\ell + \mathbf{S}_\ell) & \quad \textit{otherwise}
        \end{array}
    \right. \\
    &\mathbf{E}_\ell &&= \textit{LayerNorm}(\textit{FeedForward}(\mathbf{T}_\ell)) + \mathbf{T}_\ell
\end{alignat*}
The intermediate outputs of the attention stack are $\mathbf{S}^a_\ell \Rightarrow \mathbf{S}^b_\ell \Rightarrow \mathbf{S}_\ell$. To reproduce BERT-fused, we remove $\mathbf{S}^a_\ell$ and $\mathbf{S}^b_\ell$ from the stack and set $\mathbf{S}_\ell$ directly to $\textit{AttnBlock}(\mathbf{A},\ \mathbf{A},\ \mathbf{E}_{\ell - 1})$. We use \textit{attention block} to refer to the attention, layer normalization, and residual operations,
\begin{multline*}
    \textit{AttnBlock}(\mathbf{K}, \mathbf{V}, \mathbf{Q}) = \\
    \textit{LayerNorm}(\textit{Attn}(\mathbf{K}, \mathbf{V}, \mathbf{Q})) + \mathbf{Q}
\end{multline*}
While \textit{drop-branch} \cite{fan2020multibranch} is defined as
\begin{multline*}
    \textit{DropBranch}(\mathbf{M}, \mathbf{N}) = \\
    \mathbbm{1}(\mathbf{u} \geq .5) \cdot \mathbf{M} + \mathbbm{1}(\mathbf{u} < .5) \cdot \mathbf{N}
\end{multline*}
where $\mathbf{u} \sim \mathrm{Uniform}(0, 1)$ and $\mathbbm{1}$ is the indicator function.

\section{Experiment Setup}

\subsection{Datasets}
We evaluate our model on three translation tasks, the NIST Open MT Chinese--English task,\footnote{\url{https://www.nist.gov/itl/iad/mig/open-machine-translation-evaluation}} the IWSLT'14 English-German translation task,\footnote{\url{https://sites.google.com/site/iwsltevaluation2014/mt-track}} and the WMT'14 English-German news translation task.\footnote{\url{http://statmt.org/wmt14/translation-task.html}} Table \ref{table:training_data_quantity} provides a breakdown of the type, quantity, and relevance of the data used in the various dataset treatments. NIST provides the largest amount of in domain contextualized sentence pairs. IWSLT'14 and WMT'14 are almost an order of magnitude smaller. See Appendix A for preprocessing details.

\paragraph{NIST} Chinese--English is comprised of LDC distributed news articles and broadcast transcripts. We use the MT06 dataset as validation set and MT03, MT04, MT05, and MT08 as test sets. The validation set contains 1,649 sentences and the test set 5,146 sentences. Chinese sentences are frequently underspecified with respect to grammatical features that are obligatory in English (e.g., number for nouns, tense on verbs, and dropped arguments), making it a common language pair to study for document translation.

\paragraph{IWSLT'14} English--German is a corpus of translated TED and TEDx talks. Following prior work \citep{bert-mt}, we use the combination of {\it dev2010}, {\it dev2012}, {\it tst2010}, {\it tst2011}, and {\it tst2012} as the test set which contains 6,750 sentences. We randomly selected 10 documents from the training data for validation. We perform a data augmentation experiment with this dataset by additionally including \textit{news commentary v15}. We denote this treatment as IWSLT+ and consider this to be out of domain data augmentation.

\paragraph{WMT'14} English--German is a collection of web data, news commentary, and news articles. We use {\it newstest2013} for validation and {\it newstest2014} as the test set. For the document data, we use the original WMT'14 \textit{news commentary v9} dataset. We run two document augmentation experiments on this dataset. The first, denoted as WMT+, replaces \textit{news commentary v9} with the newer \textit{news commentary v15} dataset. The second augmentation experiment, denoted as WMT++, builds on the first by additionally incorporating the Tilde Rapid 2019 corpus. The Rapid corpus is comprised of European Commission press releases and the language style is quite different from the style used in the News Commentary data. For this reason, we consider Rapid to be out of domain data for this task.

\begin{table}
\centering
\begin{tabular}{l|rr|rr}
\hline
% \multicolumn{1}{c|}{\textbf{Dataset}} & \multicolumn{2}{c|}{\textbf{Sentence}} & \multicolumn{2}{c}{\textbf{Document}} \\
\multicolumn{1}{c|}{\textbf{Dataset}} & \multicolumn{2}{|c|}{\textbf{In Domain}} & \multicolumn{2}{|c}{\textbf{Out Domain}} \\
& \multicolumn{1}{|c}{Sent} & \multicolumn{1}{c|}{Doc} & \multicolumn{1}{|c}{Sent} & \multicolumn{1}{c}{Doc} \\ 
\hline
NIST & 1.45M & 1.45M & \multicolumn{1}{c}{-} & \multicolumn{1}{c}{-} \\
\hline
IWSLT & 173K & 173K & \multicolumn{1}{c}{-} & \multicolumn{1}{c}{-} \\
IWSLT+ & 173K & 173K & 345K & 345K \\
\hline
WMT & 4.7M & 200K & \multicolumn{1}{c}{-} & \multicolumn{1}{c}{-} \\
WMT+ & 4.85M & 345K & \multicolumn{1}{c}{-} & \multicolumn{1}{c}{-} \\
WMT++ & 4.85M & 345K & 1.63M  & 1.63M \\
\hline
\end{tabular}
\caption{\label{table:training_data_quantity} We breakdown the type, quantity, and relevance of parallel sentences used when training models for each dataset. Taking into account input requirements, models were trained on the sum of the in domain and out of domain data for a given dataset treatment. The ratio of in domain vs out of domain data per training batch was tuned on the validation set for each treatment. We used the dataset descriptions to determine the domain. For example, IWSLT'14 is a dataset of translated TED talks so we considered News Commentary data which is composed of translated news articles to be out of domain for this task.}
\end{table}

\begin{table*}[t]
\centering
\begin{tabular}{llllccc}
\hline 
& &  &  & Zh\textbar En & En\textbar De & En\textbar De \\ 
\textbf{} & \textbf{Model} & \textbf{Type} & \textbf{Embeddings} & \textbf{NIST} & \textbf{IWSLT} & \textbf{WMT} \\ 
\hline
\multirow{3}{*}{\shortstack[l]{Base-\\lines}} & Transformer & sent  & - & 46.69 & 28.68 & 28.46 \\
  & Doc Transformer & doc & $\textit{D}_\textit{s}\textit{(p,c)}$ & 47.28 & 28.74 & - \\
  & BERT-fused & doc & $\model{B}{s}{p,c}$ & 50.08 & 29.44 & 28.35 \\
\hline
\multirow{3}{*}{\shortstack[l]{This \\ work}} & Multi-source & sent & $\model{B}{s}{c} \Rightarrow \model{P}{s}{c}$ & 49.72 & \textbf{30.17} & \textbf{29.65} \\
& Multi-context & doc & $\model{B}{s}{p,c} \Rightarrow \model{B}{s}{c,n} \Rightarrow \model{P}{s}{3p,c,3n}$ & \textbf{51.07} & 29.97 & 28.11 \\
& $\quad \quad \ $+ target & doc & $\textit{Multi-context} \Rightarrow \model{P}{t}{3p,3n}$ & 50.93 & 30.10 & 28.26 \\
\hline
\end{tabular}
\caption{\label{table:main_results} Our two main findings, sacreBLEU on \textbf{Test}. (1) Source embedding enrichment, represented by our Multi-source model, provides a substantial boost to the baseline transformer model. (2) With adequate quantities of paired document training data, models that incorporate extra-sentential context provide an additional performance gain. }
\end{table*}

\subsection{Training}

We construct enhanced models with additional attention blocks and restore all previously trained parameters. We randomly initialize the newly added parameters and exclusively update these during training. For a given dataset, we train a model on all the training data it is compatible with. This means that for document-level models, only document data is used, while for sentence-level models both document and sentence data is used. In our work, this distinction only matters for the WMT'14 dataset where there is a large disparity between the two types of data.

Transformer models are trained on sentence pair data to convergence. For NIST and IWSLT'14 we use \textit{transformer base} while for WMT'14 we use \textit{transformer big}. We use the following variants of BERT from Google Research GitHub:\footnote{\url{https://github.com/google-research/bert}} BERT-Base Chinese on NIST, BERT-Base Uncased on IWSLT'14, and BERT-Large Uncased (Whole Word Masking) on WMT'14. We pretrain three PEGASUS base models for the languages en, de, and zh using the Multilingual C4 dataset as detailed in TensorFlow's dataset catalog.\footnote{\tfdlink} When training our models, we only mask a single sentence per training example and do not include a masked word auxiliary objective. We use the public PEGASUS large\footnote{\url{https://github.com/google-research/pegasus}} on the English side of WMT'14, for everything else, we use our models. See Appendix B for batch size and compute details.

\subsection{Evaluation}

To reduce the variance of our results and help with reproducibility, we use \textit{checkpoint averaging}. We select the ten contiguous checkpoints with the highest average validation BLEU. We do this at two critical points: (1) with the transformer models used to bootstrap enhanced models; (2) before calculating the validation and test BLEU scores we report. We use the \textit{sacreBLEU} script \cite{post-2018-call}\footnote{\url{https://github.com/mjpost/sacreBLEU}} on our denormalized output to calculate BLEU.

\section{Results}

In this section, we present our main results and explore the importance of each component in the multi-context model. Additionally, we investigate the performance impact of document-level parallel data scarcity, the value of source-side versus target-side context, and the importance of target context quality. 

%\subsection{Diverse Source Embeddings}

Table \ref{table:main_results} compares our Multi-source and Multi-context models to baselines of related prior work, transformer \citep{VaswaniSPUJGKP17}, document transformer \citep{ZhangLSZXZL18}, and the BERT-fused model for machine translation \citep{bert-mt}. We see that a multi-embedding model outperforms all the single embedding models in each of the datasets we try. However, the best multi-embedding configuration varies by dataset. We find that incorporating target-side context does not improve performance beyond using source-side context alone. We will present our ablation studies in the subsequent sections to further shed light on the causes of this pattern of results. To preserve the value of test set, we report results on the validation set for these experiments.

%To better understand where the performance gain is coming from and the conditions necessary for a document-level model to outperform its sentence-level counterpart, we perform a series of ablation studies.
%We then perform a series of ablation studies in order to better understand where the performance gain is coming from and the conditions necessary for a document-level model to outperform its sentence-level counterpart.

%\section{Ablations and Analysis}

\subsection{Source Context vs. Target Context}
In some language pairs, the source language is underspecified with respect to the obligatory information that must be given in the target language. For example, in English every inflected verb must have tense and this is generally not overtly marked in Chinese. In these situations, being able to condition on prior translation decisions would be valuable. However, in practice, the target context is only available post translation, meaning there is a risk of cascading errors. In this section, we seek to answer two questions: (1)~how does the quality of target context affect document-level translation; (2)~whether incorporating high-quality target context into source only models adds additional value. 

To answer the first question, we evaluate the target context model $\model{P}{t}{3p,3n}$ using various translations as context. Table  \ref{table:target_context_quality} shows the BLEU scores achieved by the target context models on the validation set. The lowest quality context comes from using the output of the baseline transformer model to furnish the context (valid BLEU of 48.76); the middle level comes from a model that conditions on three views of source context (valid BLEU of 52.8) and the third is an oracle experiment that uses a human reference translation. We see that the BLEU score improves as the quality of the target context improves; however, the impact is still less than the Multi-context source model---even in the oracle case!

\begin{table}
\centering
\begin{tabular}{llc}
\multicolumn{3}{c}{NIST Zh $\rightarrow$ En} \\
\multicolumn{3}{c}{Target Context Quality} \\
\hline
\textbf{Model} & \textbf{Context Quality} $\uparrow$ & \textbf{Valid}\\ \hline
Transformer & - & 48.76 \\
%$\model{P}{s}{3p,3n}$ & - & 49.46 \\
\hline
\multirow{3}{*}{$\model{P}{t}{3p,3n}$} & 48.76 & 49.35 \\
& 52.80 & 49.83 \\
& 100.00 & 50.32 \\
\hline
\end{tabular}
\caption{\label{table:target_context_quality} The value of using context on the target side of a translation is dependent on its quality. We test this in the limit by providing oracle context, which uses one of the references as context. We report BLEU scores on the validation set. The numbers in the second column are the BLEU scores of the translations used as the context, indicating the quality of the context.}
\end{table}

Next, we explore whether leveraging both source and target context works better than only using source context. To control for the confounding factor of target context quality, we remove one of the references from the validation dataset and use it only as context. We believe this provides an upper bound on the effect of target context for two reasons: (1) it's reasonable to assume that, at some point, machine translation will be capable of generating human quality translations; (2) even when this occurs, we will not have access to the style of a specific translator ahead of time. For these reasons, we calculate BLEU scores using only the three remaining references. We can see in Table \ref{table:target_context_ablation} that adding human quality target context to Multi-context only produces a 0.14 BLEU improvement. This challenges the notion that target context can add more value than source context alone.

 %We train models that leverage both the source and target context to determine how useful target context is. To evaluate these models we remove one of the references from the validation dataset and use it as context only. We use the remining three references to calculate BLEU. We can see in Table \ref{table:target_context_ablation} that the difference between Multi-context and Multi-context $\model{P}{t}{3p,3n}$ is only .14 BLEU. This challenges the notion that target context can add more value than source context alone.

\begin{table}
\centering
\begin{tabular}{llc}
\multicolumn{3}{c}{NIST Zh $\rightarrow$ En} \\
\multicolumn{3}{c}{Two Sided Context} \\
\hline 
\textbf{Side} & \textbf{Model} & \textbf{Valid} \\ \hline
 & Transformer & 42.51 \\
tgt & $\model{P}{t}{3p,3n}$ & 43.51 \\
src & Multi-source & 44.42 \\
src & Multi-context & 45.93 \\
both & Multi-context $\Rightarrow \model{P}{t}{3p,3n}$ & 46.07 \\
\hline
\end{tabular}
\caption{\label{table:target_context_ablation} We remove one of the references from the validation dataset and use it to provide target context only. The numbers are lower compared to other tables because the BLEU score is calculated w.r.t three references instead of four. Using human level target context offers little value over using source context alone.}
\end{table}

\subsection{Context Ablation}
\label{sec:context_ablation}

To assess the importance of the various embeddings incorporated in the Multi-context model, we perform an ablation study by adding one component at a time until we reach its full complexity. Table \ref{table:source_context_ablation} shows the study results. We can see that much of the improvement comes from the stronger sentence-level model produced by adding BERT's encoding of the source sentence---a full 2.25 BLEU improvement. The benefit of providing contextual embeddings is more incremental, yet consistent. Adding the previous sentence gives us 0.44 BLEU, adding additional depth provides another .49, and including the next sentence adds .37. Finally, adding PEGASUS' contextual embedding on top of all this results in a boost of .49. Holistically, we can assign 2.45 BLEU to source embedding enrichment and 1.59 to contextual representations.

\begin{table}
\centering
\begin{tabular}{lc}
\multicolumn{2}{c}{NIST Zh $\rightarrow$ En} \\
\multicolumn{2}{c}{Embedding Ablation} \\
\hline 
\textbf{Embeddings} & \textbf{Valid} \\ \hline
Transformer & 48.76 \\
$\model{B}{s}{c}$ & 51.01 \\
$\model{B}{s}{c} \Rightarrow \model{P}{s}{c}$ & 51.21 \\
\hline
% $\textit{D}_\textit{s}\textit{(p,c)}$ & 49.15 \\
$\model{B}{s}{p,c}$ & 51.45 \\ 
{\color{gray} $\model{B}{s}{p,c} \Rightarrow \model{B}{s}{p,c}$} & {\color{gray}51.94} \\
$\model{B}{s}{p,c} \Rightarrow \model{B}{s}{c,n}$ & 52.31 \\
$\model{B}{s}{p,c} \Rightarrow \model{P}{s}{3p,c,3n}$ & 52.30 \\
{\color{gray}$\model{B}{s}{p,c} \Rightarrow \model{B}{s}{p,c} \Rightarrow \model{B}{s}{c,n}$} & {\color{gray}52.10} \\
$\model{B}{s}{p,c} \Rightarrow \model{B}{s}{c,n} \Rightarrow \model{P}{s}{3p,c,3n}$ & \textbf{52.80} \\
\hline
\end{tabular}
\caption{\label{table:source_context_ablation} We perform ablation experiments on the NIST validation dataset to better understand the performance increase of the Multi-context model. We conclude that, in this document rich environment, multiple sources of embedding enrichment and document context contribute to performance. Adding additional parameters also helps but we only see this when going from one to two blocks. Parameter control experiments are shown in light grey.}
\end{table}

\subsection{Data Scarcity}
\label{sec:data_ablation}

\begin{figure*}[t]
\centering
Impact of Data Scarcity\par
\includegraphics[width=1.0\textwidth]{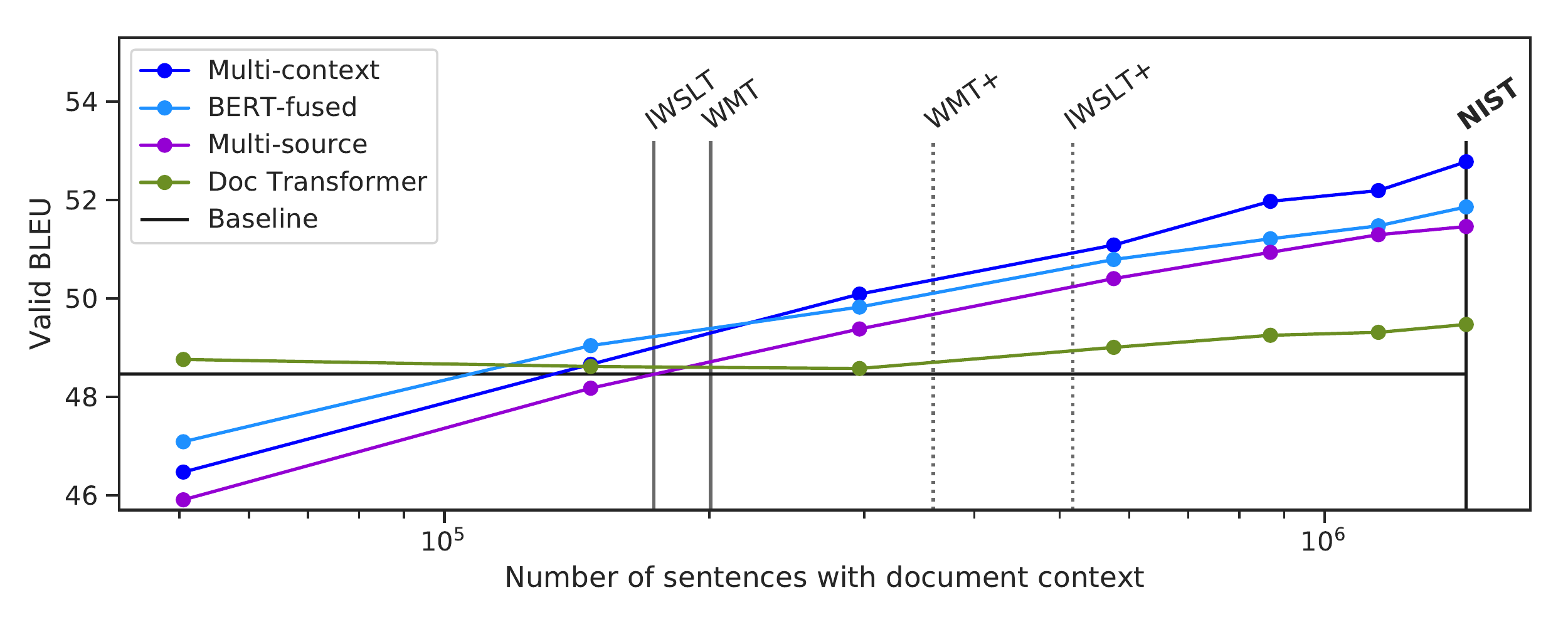}
\caption{Document context models require sufficient contextualized training data in order to be effective. We simulate data scarcity on the NIST dataset by randomly sampling a subset of the data and using it to train the various models. In order to outperform the baseline, pretrained models need 300k examples while the Doc Transformer needs 600K examples. }
\label{fig:data_scarcity}
\end{figure*}

NIST is a high resource document dataset containing over 1.4M contextualized sentence pairs. In this section, we investigate to what extent the quantities of parallel documents affect the performance of our models. To do so, we retrain enhanced models with subsets of the NIST training dataset. It is important to note that the underlying sentence transformer model was \emph{not} retrained in these experiments meaning that these experiments simulate adding document context to a strong baseline as done in \citet{lopes-etal-2020-document}. Figure \ref{fig:data_scarcity} shows the BLEU scores of different models on the NIST validation set with respect to the number of contextualized sentences used for training. We can see that it requires an example pool size over 300K before these models outperform the baseline. We conjecture that sufficient contextualized sentence pairs are crucial for document-level models to achieve good performance, which would explain why these models don’t perform well on the IWSLT’14 and WMT’14 datasets. 

Further, this pattern of results helps shed light on the inconsistent findings in the literature regarding the effectiveness of document context models. A few works~\citep{kim-etal-2019-document,li-etal-2020-multi-encoder,lopes-etal-2020-document} have found that the benefit provided by many document context models can be explained away by factors other than contextual conditioning. We can now see from Figure \ref{fig:data_scarcity} that these experiments were done in the low data regime. The randomly initialized context model needs around 600K training examples before it significantly outperform the baseline, while the pretrained contextual models reduce this to about 300K. It is important to note that none of the conextual models we tried outperformed the baseline below this point. This indicates that data quantity is not the only factor that matters but it is a prerequisite for the current class of document context architectures.

\subsection{Document Data Augmentation}
\label{sec:data_augmentation}

%In light of the data requirements of document context models, we experiment with document data augmentation. 
We further validate our hypothesis about the importance of sufficient contextualized data by experimenting with document data augmentation, this time drawing data from different domains.
We augment the IWSLT dataset with \textit{news commentary v15}, an additional 345K document context sentence pairs, and repeat the IWSLT experiments. During training, we sample from the datasets such that each batch contains roughly 50\% of the original IWSLT data. To ensure a fair comparison, we first finetune the baseline transformer model on the new data, which improves its performance by 1.61 BLEU. We use this stronger baseline as the foundation for the other models and show the results in Table \ref{table:doc_augmentation_results}. Although Multi-context edges ahead of Multi-source, the significance lies in the relative impact additional document data has on the two classes of models. The average improvement of the sentence-level models is 1.58 versus the 1.98 experienced by the document models. \citet{huo-etal-2020-diving} observed a similar phenomenon when using synthetic document augmentation. This further emphasizes the importance of using sufficient contextualized data when comparing the impact of various document-level architectures, even when the contextualized data is drawn from a new domain.

\begin{table}
\centering
\begin{tabular}{llcc}
\multicolumn{4}{c}{IWSLT'14 En $\rightarrow$ De} \\
\multicolumn{4}{c}{Document Augmentation} \\
\hline 
\textbf{Type} & \textbf{Model} & \textbf{IWSLT} & \textbf{IWSLT+} \\ \hline
\multirow{2}{*}{Sent}&Transformer & 28.68 & 30.29  \\
&Multi-source & \textbf{30.17}  &  31.71 \\
\hline
\multirow{2}{*}{Doc}&BERT-fused & 29.44 & 31.50 \\
&Multi-context & 29.97 & \textbf{31.86} \\
\hline
\end{tabular}
\caption{\label{table:doc_augmentation_results} Model performance before and after document data augmentation. We see that most of the improvement is coming from source embedding enrichment. Data augmentation is required for document-level models to additionally learn to leverage contextual information. The document-level models benefit significantly more from additional document data than the sentence-level models. }
\end{table}

\subsection{Three Stage Training}

WMT'14 offers an opportunity to combine the insights gained from the aforementioned experiments. This dataset provides large quantities of sentence pair data and a small amount of document pair data. Not surprisingly, both BERT-fused\footnote{Here we mention that, while we were able to reproduce the baseline relative uplift of BERT-fused on the other datasets, we were unable to do so on the WMT'14 dataset. We do not know what document data they used and this probably accounts for the differences observed.} and Multi-context struggle in this environment. On the other hand, Multi-source benefits from the abundance of sentence pair data.

In order to make the most of the training data, we add a third stage to our training regime. As before, in stage one, we train the transformer model with the sentence pair data. In stage two, we train the Multi-source model also using the sentence pair data. In stage three, we add an additional $\model{P}{s}{3p,3n}$ attention block to the Multi-source model and train it with document data. We perform two document augmentation experiments. In the first, we replace \textit{news commentary v9} with \textit{v15}. In the second, we train on a mix of \textit{news commentary v15} and Tilde Rapid 2019. The optimal mix was 70\% and 30\% respectably, which we found by tuning on the validation dataset. For each of the augmentation experiments, we created new Multi-source baselines by fine-tuning the original baseline on the new data. When training these new baselines we only updated the parameters in the $\model{B}{s}{c}$ and $\model{P}{s}{c}$ attention blocks. In contrast, when training the treatment models, we froze these blocks and only updated the parameters in the $\model{P}{s}{3p,3n}$ block. In this way, both the new baselines and treatments started from the same pretrained Multi-source model, were exposed to the same data, and had only the parameters under investigation updated. 

We see in Table \ref{table:wmt14_results} that this method can be used to provide the document-level model with a much stronger sentence-level model to start from. As we saw in the previous data augmentation experiments (\S\ref{sec:data_augmentation}), document augmentation helps the document-level model more than the sentence-level model. It is interesting to note that out of domain document data helps the document-level model yet hurts the sentence-level model.\footnote{While tuning on the validation dataset, we observed that the optimal proportion of Rapid data to include for the new baseline was 0\%. Meaning, don't include any of the off domain data. However, we needed a fair comparison baseline so left it at 30\% when making Table \ref{table:wmt14_results}.}

\begin{table}
\centering
\begin{tabular}{c|llc}
\multicolumn{4}{c}{WMT'14 En $\rightarrow$ De} \\
\multicolumn{4}{c}{Three Stage Training} \\
\hline 
\textbf{Stage} & \textbf{Model}  & \textbf{Data} & \textbf{Test} \\ \hline
1 & Transformer & sent & 28.46 \\
\hline 
2 & Multi-source & sent-WMT & 29.64 \\
\hline
\multirow{5}{*}{3} & \multirow{2}{*}{Multi-source} & sent-WMT+ & 29.74 \\
& & sent-WMT++ & 29.62 \\
\cline{2-4}
& \multirow{3}{2.5cm}{Multi-source \\ $\quad \Rightarrow \model{P}{s}{3p,3n}$} & doc-WMT & 29.60 \\
& & doc-WMT+ & 29.78 \\
& & doc-WMT++ & \textbf{29.89} \\
\hline
\end{tabular}
\caption{\label{table:wmt14_results} Results from using a three staged training approach. When there is large disparity between the amount of sentence pair data and document data, this method enables training new attention blocks with the maximum amount of available data given their input restrictions.   }
\end{table}

% \subsection{Embedding Attention Architecture}
% \lei{djd: consider moving this section to Section 2.3/2.4. You don't need the table here, just say preliminary experiments show that the parallel architecture perform the best and we stick to this architecture in all the experiments.}
% There are a few ways to integrate the output of external models into a transformer layer. We could stack them \textit{vertically} after the self-attention block \cite{ZhangLSZXZL18} or we could place them \textit{horizontally} and average all of their outputs together like MAT \cite{fan2020multibranch}. We compare these alternative configurations to our parallel stack arrangement and find that there is tangible benefit to using our style. It's important to point out that MAT was not designed to incorporate external embeddings as we do here, but is a reasonable alternative configuration. Table \ref{table:source_architecture_ablation} shows the results of this ablation on the Multi-context model.

% \begin{table}
% \centering
% \begin{tabular}{llc}
% \multicolumn{3}{c}{NIST Zh $\rightarrow$ En} \\
% \multicolumn{3}{c}{Architecture Comparison} \\
% \hline
% \textbf{Model} & \textbf{Architecture} & \textbf{BLEU} \\ \hline
% \multirow{3}{*}{Multi-context} & Horizontal & 51.95 \\
% & Vertical & 52.20 \\
% & Parallel & \textbf{52.80} \\
% \hline
% \end{tabular}
% \caption{\label{table:source_architecture_ablation} Architecture ablation using the NIST validation dataset on the Multi-context model.  }
% \end{table}

\section{Related Work}
This work is closely related to two lines of research: document-level neural machine translation and representation learning via language modeling.

Earlier work in document machine translation exploits the context by taking a concatenated string of adjacent source sentences as the input of neural sequence-to-sequence models \citep{TiedemannS17}. Follow-up work adds additional context layers to the neural sequence-to-sequence models in order to have a better encoding of the context information \citep[\textit{inter alia}]{ZhangLSZXZL18,WerlenRPH18}. They vary in terms of whether to incorporate the source-side context \citep{BawdenSBH18,ZhangLSZXZL18,WerlenRPH18} or target-side context \citep{TuLSZ18}, and whether to condition on a few adjacent sentences  \citep{JeanLFC17,WangTWL17,TuLSZ18,TitovSSV18,ZhangLSZXZL18,WerlenRPH18} or the full document \citep{HaffariM18,MarufMH19}. Our work is similar to this line of research since we have also introduced additional attention components to the transformer. However, our model is different from theirs in that the context encoders were pretrained with a masked language model objective. 

There has also been work on leveraging monolingual documents to improve document-level machine translation. \citet{Junczys-Dowmunt19} creates synthetic parallel documents generated by backtranslation \citep{DBLP:conf/acl/SennrichHB16,EdunovOAG18} and uses the combination of the original and the synthetic parallel documents to train the document translation models. \citet{voita-2019-context} train a post-editing model from monolingual documents to post-edit sentence-level translations into document-level translations. \citet{DBLP:journals/tacl/YuSSLKBD20,DBLP:conf/wmt/YuSHSDSALMLDYBD20} uses Bayes' rule to combine a monolingual document language model probability with sentence translation probabilities.
%\citet{DBLP:journals/tacl/YuSSLKBD20} and  \citet{DBLP:conf/wmt/YuSHSDSALMLDYBD20} use a large-scale language model for scoring in Bayes' rule derived document-level translation models. Different from their work, our model integrates context encoders pretrained with monolingual documents. 

Finally, large-scale representation learning with language modeling has achieved success in improving systems in language understanding, leading to state-of-the-art results on a wide range of tasks \citep{Peters:2018,devlin-2019-bert,radford2018improving,DBLP:conf/nips/McCannBXS17,DBLP:journals/corr/abs-1906-08237,DBLP:conf/naacl/ChronopoulouBP19,DBLP:journals/corr/abs-1901-07291,gpt-3}. They have also been used to improve text generation tasks, such as sentence-level machine translation \citep{DBLP:conf/icml/SongTQLL19,DBLP:conf/naacl/EdunovBA19,bert-mt} and summarization \citep{DBLP:journals/corr/abs-1902-09243,pegasus,DBLP:journals/corr/abs-1905-03197}, and repurposing unconditional language generation \citep{ziegler:2019,Oliveira2019RepurposingDL}. Our work is closely related to that from \citet{bert-mt}, where pretrained large-scale language models are applied to document-level machine translation tasks. We advance this line of reasoning by designing an architecture that uses composition to incorporate \textit{multiple} pretrained models at once. It also enables conditioning on different inputs to the same pretrained model, enabling us to circumvent BERT's two sentence embedding limit.

%\citep{Peters:2018,DBLP:conf/naacl/DevlinCLT19,radford2018improving,DBLP:conf/nips/McCannBXS17,DBLP:journals/corr/abs-1906-08237,DBLP:conf/naacl/ChronopoulouBP19,DBLP:journals/corr/abs-1901-07291}. Language generation is another area where pretrained language models have been applied, with existing work focusing on fine-tuning for repurposing an unconditional language model %\citep{DBLP:journals/corr/abs-1902-09243,DBLP:conf/naacl/EdunovBA19,DBLP:conf/icml/SongTQLL19,DBLP:journals/corr/abs-1905-03197,ziegler:2019,Oliveira2019RepurposingDL}. 

\section{Conclusion}

We have introduced an architecture and training regimen that enables incorporating representations from multiple pretrained masked language models into a transformer model. We show that this technique can be used to create a substantially stronger sentence-level model and, with sufficient document data, further upgraded to a document-level model that conditions on contextual information. Through ablations and other experiments, we establish document augmentation and multi-stage training as effective strategies for training a document-level model when faced with data scarcity. And that source side context is sufficient for these models, with target context adding little additional value.

\section*{Acknowledgments}
We would like to thank our teammates, Laurent Sartran, Phil Blunsom, Susie Young, Wang Ling, and Wojciech Stokowiec, for their feedback and shared engineering efforts. We thank Yao Zhao for helping us to better understand the PEGASUS codebase. We thank Dani Yogatama and our three anonymous reviewers for their feedback on the earlier draft of the paper. Their feedback was taken seriously and we believe this work has benefited from the items they requested.
% The acknowledgments should go immediately before the references. Do not number the acknowledgments section.
% \textbf{Do not include this section when submitting your paper for review.}
\bibliographystyle{acl_natbib}
\bibliography{acl2021}

\begin{thebibliography}{44}
\expandafter\ifx\csname natexlab\endcsname\relax\def\natexlab#1{#1}\fi

\bibitem[{Bahdanau et~al.(2015)Bahdanau, Cho, and Bengio}]{bahdanau:2015}
Dzmitry Bahdanau, Kyunghyun Cho, and Yoshua Bengio. 2015.
\newblock Neural machine translation by jointly learning to align and
  translate.
\newblock In \emph{Proceedings of ICLR}.

\bibitem[{Bawden et~al.(2018)Bawden, Sennrich, Birch, and Haddow}]{BawdenSBH18}
Rachel Bawden, Rico Sennrich, Alexandra Birch, and Barry Haddow. 2018.
\newblock Evaluating discourse phenomena in neural machine translation.
\newblock In \emph{Proceedings of {NAACL-HLT}}.

\bibitem[{Brown et~al.(2020)Brown, Mann, Ryder, Subbiah, Kaplan, Dhariwal,
  Neelakantan, Shyam, Sastry, Askell, Agarwal, Herbert{-}Voss, Krueger,
  Henighan, Child, Ramesh, Ziegler, Wu, Winter, Hesse, Chen, Sigler, Litwin,
  Gray, Chess, Clark, Berner, McCandlish, Radford, Sutskever, and
  Amodei}]{gpt-3}
Tom~B. Brown, Benjamin Mann, Nick Ryder, Melanie Subbiah, Jared Kaplan,
  Prafulla Dhariwal, Arvind Neelakantan, Pranav Shyam, Girish Sastry, Amanda
  Askell, Sandhini Agarwal, Ariel Herbert{-}Voss, Gretchen Krueger, Tom
  Henighan, Rewon Child, Aditya Ramesh, Daniel~M. Ziegler, Jeffrey Wu, Clemens
  Winter, Christopher Hesse, Mark Chen, Eric Sigler, Mateusz Litwin, Scott
  Gray, Benjamin Chess, Jack Clark, Christopher Berner, Sam McCandlish, Alec
  Radford, Ilya Sutskever, and Dario Amodei. 2020.
\newblock Language models are few-shot learners.
\newblock In \emph{Proceedings of {NeurIPS}}.

\bibitem[{Chen et~al.(2018)Chen, Firat, Bapna, Johnson, Macherey, Foster,
  Jones, Schuster, Shazeer, Parmar, Vaswani, Uszkoreit, Kaiser, Chen, Wu, and
  Hughes}]{chen-etal-2018-best}
Mia~Xu Chen, Orhan Firat, Ankur Bapna, Melvin Johnson, Wolfgang Macherey,
  George Foster, Llion Jones, Mike Schuster, Noam Shazeer, Niki Parmar, Ashish
  Vaswani, Jakob Uszkoreit, Lukasz Kaiser, Zhifeng Chen, Yonghui Wu, and
  Macduff Hughes. 2018.
\newblock The best of both worlds: Combining recent advances in neural machine
  translation.
\newblock In \emph{Proceedings of ACL}.

\bibitem[{Chronopoulou et~al.(2019)Chronopoulou, Baziotis, and
  Potamianos}]{DBLP:conf/naacl/ChronopoulouBP19}
Alexandra Chronopoulou, Christos Baziotis, and Alexandros Potamianos. 2019.
\newblock An embarrassingly simple approach for transfer learning from
  pretrained language models.
\newblock In \emph{Proceedings of {NAACL-HLT}}.

\bibitem[{Devlin et~al.(2019)Devlin, Chang, Lee, and
  Toutanova}]{devlin-2019-bert}
Jacob Devlin, Ming-Wei Chang, Kenton Lee, and Kristina Toutanova. 2019.
\newblock {BERT: Pre-training of Deep Bidirectional Transformers for Language
  Understanding}.
\newblock In \emph{Proceedings of {NAACL-HLT}}.

\bibitem[{Dong et~al.(2019)Dong, Yang, Wang, Wei, Liu, Wang, Gao, Zhou, and
  Hon}]{DBLP:journals/corr/abs-1905-03197}
Li~Dong, Nan Yang, Wenhui Wang, Furu Wei, Xiaodong Liu, Yu~Wang, Jianfeng Gao,
  Ming Zhou, and Hsiao{-}Wuen Hon. 2019.
\newblock Unified language model pre-training for natural language
  understanding and generation.
\newblock \emph{CoRR}, abs/1905.03197.

\bibitem[{Edunov et~al.(2019)Edunov, Baevski, and
  Auli}]{DBLP:conf/naacl/EdunovBA19}
Sergey Edunov, Alexei Baevski, and Michael Auli. 2019.
\newblock Pre-trained language model representations for language generation.
\newblock In \emph{Proceedings of {NAACL-HLT}}.

\bibitem[{Edunov et~al.(2018)Edunov, Ott, Auli, and Grangier}]{EdunovOAG18}
Sergey Edunov, Myle Ott, Michael Auli, and David Grangier. 2018.
\newblock Understanding back-translation at scale.
\newblock In \emph{Proceedings of {EMNLP}}.

\bibitem[{Fan et~al.(2020)Fan, Xie, Xia, Wu, Qin, Li, and
  Liu}]{fan2020multibranch}
Yang Fan, Shufang Xie, Yingce Xia, Lijun Wu, Tao Qin, Xiang-Yang Li, and
  Tie-Yan Liu. 2020.
\newblock \href {http://arxiv.org/abs/2006.10270} {Multi-branch attentive
  transformer}.

\bibitem[{Gamma et~al.(1995)Gamma, Helm, Johnson, and
  Vlissides}]{designPatterns}
Erich Gamma, Richard Helm, Ralph Johnson, and John Vlissides. 1995.
\newblock \emph{Design Patterns: Elements of Reusable Object-Oriented
  Software}.
\newblock Addison-Wesley Longman Publishing Co., Inc., USA.

\bibitem[{Haffari and Maruf(2018)}]{HaffariM18}
Gholamreza Haffari and Sameen Maruf. 2018.
\newblock Document context neural machine translation with memory networks.
\newblock In \emph{Proceedings of {ACL}}.

\bibitem[{Huo et~al.(2020)Huo, Herold, Gao, Dahlmann, Khadivi, and
  Ney}]{huo-etal-2020-diving}
Jingjing Huo, Christian Herold, Yingbo Gao, Leonard Dahlmann, Shahram Khadivi,
  and Hermann Ney. 2020.
\newblock Diving deep into context-aware neural machine translation.
\newblock In \emph{Proceedings of {WMT}}.

\bibitem[{Jean et~al.(2017)Jean, Lauly, Firat, and Cho}]{JeanLFC17}
S{\'{e}}bastien Jean, Stanislas Lauly, Orhan Firat, and Kyunghyun Cho. 2017.
\newblock Does neural machine translation benefit from larger context?
\newblock \emph{CoRR}, abs/1704.05135.

\bibitem[{Junczys{-}Dowmunt(2019)}]{Junczys-Dowmunt19}
Marcin Junczys{-}Dowmunt. 2019.
\newblock Microsoft translator at {WMT} 2019: Towards large-scale
  document-level neural machine translation.
\newblock In \emph{Proceedings of {WMT}}.

\bibitem[{Kim et~al.(2019)Kim, Tran, and Ney}]{kim-etal-2019-document}
Yunsu Kim, Duc~Thanh Tran, and Hermann Ney. 2019.
\newblock When and why is document-level context useful in neural machine
  translation?
\newblock In \emph{Proceedings of the Fourth Workshop on Discourse in Machine
  Translation (DiscoMT 2019)}.

\bibitem[{Kudo(2018)}]{kudo-2018-subword}
Taku Kudo. 2018.
\newblock Subword regularization: Improving neural network translation models
  with multiple subword candidates.
\newblock In \emph{Proceedings of {ACL}}.

\bibitem[{Kudo and Richardson(2018)}]{sentencepiece}
Taku Kudo and John Richardson. 2018.
\newblock {S}entence{P}iece: A simple and language independent subword
  tokenizer and detokenizer for neural text processing.
\newblock In \emph{Proceedings of {EMNLP}}.

\bibitem[{Lample and Conneau(2019)}]{DBLP:journals/corr/abs-1901-07291}
Guillaume Lample and Alexis Conneau. 2019.
\newblock Cross-lingual language model pretraining.
\newblock \emph{CoRR}, abs/1901.07291.

\bibitem[{Li et~al.(2020)Li, Liu, Wang, Jiang, Xiao, Zhu, Liu, and
  Li}]{li-etal-2020-multi-encoder}
Bei Li, Hui Liu, Ziyang Wang, Yufan Jiang, Tong Xiao, Jingbo Zhu, Tongran Liu,
  and Changliang Li. 2020.
\newblock Does multi-encoder help? a case study on context-aware neural machine
  translation.
\newblock In \emph{Proceedings of {ACL}}.

\bibitem[{Lopes et~al.(2020)Lopes, Farajian, Bawden, Zhang, and
  Martins}]{lopes-etal-2020-document}
Ant{\'o}nio Lopes, M.~Amin Farajian, Rachel Bawden, Michael Zhang, and
  Andr{\'e} F.~T. Martins. 2020.
\newblock Document-level neural {MT}: A systematic comparison.
\newblock In \emph{Proceedings of the 22nd Annual Conference of the European
  Association for Machine Translation}.

\bibitem[{Maruf et~al.(2019)Maruf, Martins, and Haffari}]{MarufMH19}
Sameen Maruf, Andr{\'{e}} F.~T. Martins, and Gholamreza Haffari. 2019.
\newblock Selective attention for context-aware neural machine translation.
\newblock In \emph{Proceedings of {NAACL-HLT}}.

\bibitem[{McCann et~al.(2017)McCann, Bradbury, Xiong, and
  Socher}]{DBLP:conf/nips/McCannBXS17}
Bryan McCann, James Bradbury, Caiming Xiong, and Richard Socher. 2017.
\newblock Learned in translation: Contextualized word vectors.
\newblock In \emph{Proceedings of {NeurIPS}}.

\bibitem[{Miculicich et~al.(2018)Miculicich, Ram, Pappas, and
  Henderson}]{WerlenRPH18}
Lesly Miculicich, Dhananjay Ram, Nikolaos Pappas, and James Henderson. 2018.
\newblock Document-level neural machine translation with hierarchical attention
  networks.
\newblock In \emph{Proceedings of {EMNLP}}.

\bibitem[{de~Oliveira and Rodrigo(2019)}]{Oliveira2019RepurposingDL}
Luke de~Oliveira and Alfredo~L\'{a}inez Rodrigo. 2019.
\newblock Repurposing decoder-transformer language models for abstractive
  summarization.
\newblock \emph{ArXiv}, abs/1909.00325.

\bibitem[{Peters et~al.(2018)Peters, Neumann, Iyyer, Gardner, Clark, Lee, and
  Zettlemoyer}]{Peters:2018}
Matthew~E. Peters, Mark Neumann, Mohit Iyyer, Matt Gardner, Christopher Clark,
  Kenton Lee, and Luke Zettlemoyer. 2018.
\newblock Deep contextualized word representations.
\newblock In \emph{Proceedings of NAACL}.

\bibitem[{Post(2018)}]{post-2018-call}
Matt Post. 2018.
\newblock A call for clarity in reporting {BLEU} scores.
\newblock In \emph{Proceedings of {WMT}}.

\bibitem[{Radford et~al.(2018)Radford, Narasimhan, Salimans, and
  Sutskever}]{radford2018improving}
Alec Radford, Karthik Narasimhan, Tim Salimans, and Ilya Sutskever. 2018.
\newblock Improving language understanding by generative pre-training.

\bibitem[{Sennrich et~al.(2016)Sennrich, Haddow, and
  Birch}]{DBLP:conf/acl/SennrichHB16}
Rico Sennrich, Barry Haddow, and Alexandra Birch. 2016.
\newblock Improving neural machine translation models with monolingual data.
\newblock In \emph{Proceedings of {ACL}}.

\bibitem[{Song et~al.(2019)Song, Tan, Qin, Lu, and
  Liu}]{DBLP:conf/icml/SongTQLL19}
Kaitao Song, Xu~Tan, Tao Qin, Jianfeng Lu, and Tie{-}Yan Liu. 2019.
\newblock {MASS:} masked sequence to sequence pre-training for language
  generation.
\newblock In \emph{Proceedings of {ICML}}.

\bibitem[{Tiedemann and Scherrer(2017)}]{TiedemannS17}
J{\"{o}}rg Tiedemann and Yves Scherrer. 2017.
\newblock Neural machine translation with extended context.
\newblock In \emph{Proceedings of DiscoMT@EMNLP}.

\bibitem[{Tu et~al.(2018)Tu, Liu, Shi, and Zhang}]{TuLSZ18}
Zhaopeng Tu, Yang Liu, Shuming Shi, and Tong Zhang. 2018.
\newblock Learning to remember translation history with a continuous cache.
\newblock \emph{{TACL}}, 6:407--420.

\bibitem[{Vaswani et~al.(2017)Vaswani, Shazeer, Parmar, Uszkoreit, Jones,
  Gomez, Kaiser, and Polosukhin}]{VaswaniSPUJGKP17}
Ashish Vaswani, Noam Shazeer, Niki Parmar, Jakob Uszkoreit, Llion Jones,
  Aidan~N. Gomez, \L{}ukasz Kaiser, and Illia Polosukhin. 2017.
\newblock Attention is all you need.
\newblock In \emph{Proceedings of {NeurIPS}}.

\bibitem[{Voita et~al.(2019)Voita, Sennrich, and Titov}]{voita-2019-context}
Elena Voita, Rico Sennrich, and Ivan Titov. 2019.
\newblock Context-aware monolingual repair for neural machine translation.
\newblock In \emph{Proceedings of {EMNLP-IJCNLP}}.

\bibitem[{Voita et~al.(2018)Voita, Serdyukov, Sennrich, and and}]{TitovSSV18}
Elena Voita, Pavel Serdyukov, Rico Sennrich, and Ivan~Titov and. 2018.
\newblock Context-aware neural machine translation learns anaphora resolution.
\newblock In \emph{Proceedings of {ACL}}.

\bibitem[{Wang et~al.(2017)Wang, Tu, Way, and Liu}]{WangTWL17}
Longyue Wang, Zhaopeng Tu, Andy Way, and Qun Liu. 2017.
\newblock Exploiting cross-sentence context for neural machine translation.
\newblock In \emph{Proceedings of {EMNLP}}.

\bibitem[{Yang et~al.(2019)Yang, Dai, Yang, Carbonell, Salakhutdinov, and
  Le}]{DBLP:journals/corr/abs-1906-08237}
Zhilin Yang, Zihang Dai, Yiming Yang, Jaime~G. Carbonell, Ruslan Salakhutdinov,
  and Quoc~V. Le. 2019.
\newblock Xlnet: Generalized autoregressive pretraining for language
  understanding.
\newblock \emph{CoRR}, abs/1906.08237.

\bibitem[{Yu et~al.(2020{\natexlab{a}})Yu, Sartran, Huang, Stokowiec, Donato,
  Srinivasan, Andreev, Ling, Mokra, Lago, Doron, Young, Blunsom, and
  Dyer}]{DBLP:conf/wmt/YuSHSDSALMLDYBD20}
Lei Yu, Laurent Sartran, Po{-}Sen Huang, Wojciech Stokowiec, Domenic Donato,
  Srivatsan Srinivasan, Alek Andreev, Wang Ling, Sona Mokra, Agustin~Dal Lago,
  Yotam Doron, Susannah Young, Phil Blunsom, and Chris Dyer.
  2020{\natexlab{a}}.
\newblock The deepmind chinese-english document translation system at
  {WMT2020}.
\newblock In \emph{Proceedings of WMT@EMNLP}.

\bibitem[{Yu et~al.(2020{\natexlab{b}})Yu, Sartran, Stokowiec, Ling, Kong,
  Blunsom, and Dyer}]{DBLP:journals/tacl/YuSSLKBD20}
Lei Yu, Laurent Sartran, Wojciech Stokowiec, Wang Ling, Lingpeng Kong, Phil
  Blunsom, and Chris Dyer. 2020{\natexlab{b}}.
\newblock Better document-level machine translation with bayes' rule.
\newblock \emph{Trans. Assoc. Comput. Linguistics}, 8:346--360.

\bibitem[{Zhang et~al.(2019)Zhang, Gong, Yan, Duan, Xu, Wang, Gong, and
  Zhou}]{DBLP:journals/corr/abs-1902-09243}
Haoyu Zhang, Yeyun Gong, Yu~Yan, Nan Duan, Jianjun Xu, Ji~Wang, Ming Gong, and
  Ming Zhou. 2019.
\newblock Pretraining-based natural language generation for text summarization.
\newblock \emph{CoRR}, abs/1902.09243.

\bibitem[{Zhang et~al.(2018)Zhang, Luan, Sun, Zhai, Xu, Zhang, and
  Liu}]{ZhangLSZXZL18}
Jiacheng Zhang, Huanbo Luan, Maosong Sun, Feifei Zhai, Jingfang Xu, Min Zhang,
  and Yang Liu. 2018.
\newblock Improving the transformer translation model with document-level
  context.
\newblock In \emph{Proceedings of {EMNLP}}.

\bibitem[{Zhang et~al.(2020)Zhang, Zhao, Saleh, and Liu}]{pegasus}
Jingqing Zhang, Yao Zhao, Mohammad Saleh, and Peter~J. Liu. 2020.
\newblock {PEGASUS:} pre-training with extracted gap-sentences for abstractive
  summarization.
\newblock In \emph{Proceedings of {ICML}}.

\bibitem[{Zhu et~al.(2020)Zhu, Xia, Wu, He, Qin, Zhou, Li, and Liu}]{bert-mt}
Jinhua Zhu, Yingce Xia, Lijun Wu, Di~He, Tao Qin, Wengang Zhou, Houqiang Li,
  and Tie{-}Yan Liu. 2020.
\newblock Incorporating {BERT} into neural machine translation.
\newblock In \emph{Proceedings of {ICLR}}.

\bibitem[{Ziegler et~al.(2019)Ziegler, Melas-Kyriazi, Gehrmann, and
  Rush}]{ziegler:2019}
Zachary~M. Ziegler, Luke Melas-Kyriazi, Sebastian Gehrmann, and Alexander~M.
  Rush. 2019.
\newblock Encoder-agnostic adaptation for conditional language generation.
\newblock \emph{CoRR}, abs/1908.06938.

\end{thebibliography}

\clearpage
\appendix
\section{Preprocessing}

\subsection{Text}
We perform text normalization on the datasets before tokenization.
\begin{itemize}
    \item All languages - Unicode canonicalization (NKFD from), replacement of common multiple encoding errors present in training data, standardization of quotation marks into ``directional'' variants.
    \item English - Replace non-American spelling variants with American spellings using the aspell library.\footnote{\url{http://wordlist.aspell.net/varcon-readme/}} Punctuation was split from English words using a purpose-built library.
    \item Chinese - Convert any traditional Chinese characters into simplified forms and segment into word-like units using the Jieba segmentation tool.\footnote{\url{https://github.com/fxsjy/jieba}}
    \item English \& German for WMT'14 - Lowercase first word of sentence unless it was in a whitelist of proper nouns and common abbreviations.
    \item English \& German for IWSLT'14 - Lowercase all words.
    \item Chinese \& English for NIST - Lowercase all words.
\end{itemize}

\subsection{Tokenization}

We encode text into sub-word units using the \texttt{sentencepiece} tool \cite{sentencepiece}. When generating our own subword segmentation, we used the algorithm from \citet{kudo-2018-subword} with a minimum character coverage of 0.9995. Other than for BERT, we use TensorFlow SentencepieceTokenizer for tokenization given a sentencepiece model.

\begin{itemize}
    \item BERT (all) - Used vocabulary provided with download and TensorFlow BertTokenizer.
    \item PEGASUS large \& EN small - Used sentencepiece model provided with PEGASUS large download.
    \item PEGASUS Zh small - Generated subword vocabulary of 34K tokens from the NIST dataset.
    \item PEGASUS De small - Generated subword vocabulary of 34K tokens from the WMT'14 dataset.
    \item Transformers - Generated joint subword vocabulary of 34K tokens for NIST \& WMT'14 and 20K for IWSLT'14.
\end{itemize}

\section{Compute}

We train and evaluate on Google TPU v2. We use a 4x2 configuration which contains 16 processing units. We use the following global batch sizes during training (examples / tokens):
\begin{itemize}
    \item Transformer baselines: (1024 / 131,072)
    \item WMT'14 Multi-source: (1024 / 131,072)
    \item WMT'14 others: (128 / 16,384)
    \item NIST: (256 / 32,767)
    \item IWSLT'14: (256 / 32,767)
\end{itemize}

Using a global batch size of 32 and a beam width of 5, the following are the number of samples per second our models and key baselines managed during inference:
\begin{itemize}
    \item Transformer: 11.94
    \item BERT-fused: 7.37
    \item Multi-source: 5.45
    \item Multi-context: 4.80
\end{itemize}

\section{Qualitative Analysis}

We manually inspected the translations outputs from the Multi-source model and Multi-context model and have found that the Multi-context model indeed does better in terms maintaining the consistency of lexical usage across sentences. Unlike English, Chinese does not mark nouns for plural vs singular nor verbs for tense. Therefore, this needs to be inferred from context to generate accurate English translations. It is not possible for a sentence-level MT system to capture this information when the relevant context is not in the current sentence. Tables \ref{example:tense_consistency},  \ref{example:proper_noun_consistency}, and \ref{example:pronoun_consistency} provide various examples where the sentence-level model cannot know this information and the document-level model is able to correctly condition on it.

\begin{CJK*}{UTF8}{gbsn}

\begin{table*}[h]
    \centering
    \renewcommand{\arraystretch}{1.2}
    \begin{tabular}{p{0.16\linewidth} p{0.75\linewidth}}
       \multicolumn{2}{c}{\textbf{Example 1}} \\
       \multicolumn{2}{c}{\textbf{Consistency of Tense}} \\
       \hline
       Source:  &  金先生说,五十几岁时,王选便开始注意培养年轻人,他一直强调,要铺路,要甘为人梯,给年轻人让路。 \\
       \hline
       Reference: & Mr. Jin said that Wang Xuan started to focus on mentoring young people when he was in his 50s. He constantly stressed that he wanted to pave the way for young people and that he wanted to be their stepping stone. \\
       \hline
       Multi-source: & Mr. Chin \colorbox{LightRed}{\textbf{says}} that when he was in his fifties, Wang began to pay attention to cultivating young people. He has always stressed that to pave the way, he must be willing to serve as a ladder for young people. \\
       \hline
       Multi-context: & Mr. Jin \colorbox{PaleGreen}{\textbf{said}} that when he was in his fifties, Wang Xuan began to pay attention to cultivating young people. He always stressed that he wanted to pave the way, to be willing to serve as a ladder, and to give young people a way. \\
       \hline
    \end{tabular}
    \caption{This came from an article describing an interview with a celebrity. The entire article used past tense.}
    \label{example:tense_consistency}
\end{table*}

\begin{table*}[h]
    \centering
    \renewcommand{\arraystretch}{1.2}
    \begin{tabular}{p{0.16\linewidth} p{0.75\linewidth}}
       \multicolumn{2}{c}{\textbf{Example 2}} \\
       \multicolumn{2}{c}{\textbf{Consistency of Proper Noun}} \\
       \hline
       Source:  &  巴政府是决不会让这种企图得逞的。 \\
       \hline
       Reference: & The Pakistani government will never allow such attempt to materialize. \\
       \hline
       Multi-source: & The \colorbox{LightRed}{\textbf{Palestinian}} government will never let this attempt succeed. \\
       \hline
       Multi-context: & The \colorbox{PaleGreen}{\textbf{Pakistani}} government will never let this attempt succeed. \\
       \hline
    \end{tabular}
    \caption{The pronoun 巴政府 \, is ambiguous since 巴 \, could be short for 巴西\,(Brazil)，巴勒斯坦\,(Palestine)，巴基斯坦\,(Pakistan). The model has to refer to the context to know that 巴 \, refers to Pakistan in this instance since this is where the entire article takes place.}
    \label{example:proper_noun_consistency}
\end{table*}

\begin{table*}[h]
    \centering
    \renewcommand{\arraystretch}{1.2}
    \begin{tabular}{p{0.16\linewidth} p{0.75\linewidth}}
       \multicolumn{2}{c}{\textbf{Example 2}} \\
       \multicolumn{2}{c}{\textbf{Consistency of Pronoun}} \\
       \hline
       Source:  &  那是在十年前的一天，当这位老师正利用中午休息时间，在家里睡觉时，突然间，电话铃响了， \\
       \hline
       Reference: & On that day ten years ago, when this teacher was taking a nap at home during noontime break, the telephone rang suddenly. \\
       \hline
       Multi-source: & That was ten years ago. When this teacher was taking advantage of \colorbox{LightRed}{\textbf{his}} lunch break, \colorbox{LightRed}{\textbf{he}} was sleeping at home. Suddenly, the phone rang. \\
       \hline
       Multi-context: & One day ten years ago, when this teacher was taking advantage of \colorbox{PaleGreen}{\textbf{her}} lunch break, \colorbox{PaleGreen}{\textbf{she}} was sleeping at home. Suddenly, the telephone rang. \\
       \hline
    \end{tabular}
    \caption{This is a story about a mother. The pronouns she/her have been used across the document. One cannot infer the gender of the teacher from the source sentence alone. Thus, the context model has to refer to the other sentences in order to get this correct.}
    \label{example:pronoun_consistency}
\end{table*}

\end{CJK*}

\end{document}